\documentclass{article}
\usepackage{spconf,graphicx,hyperref}

\usepackage{latexsym}
\usepackage{amssymb}
\usepackage{amsmath}
\usepackage{amsthm}
\usepackage{booktabs}
\usepackage{enumitem}
\usepackage{xcolor}
\usepackage{algorithm}
\usepackage{algpseudocode}
\usepackage{multirow}
\usepackage{subcaption}

\usepackage{etoolbox}
\newtoggle{arxiv}
\toggletrue{arxiv}
\title{Stabilizing PINNs: A regularization scheme for PINN training to avoid unstable fixed points of dynamical systems}
\name{Milos Babic$^{\star \bullet \ddagger}$, Franz M. Rohrhofer$^{\star}$, Bernhard C. Geiger$^{\dagger \star}$}
\address{$^{\star}$ Know Center Research GmbH, Graz, Austria\\
$^\bullet$ Christian Doppler Laboratory for Physics-driven Machine Learning in Industrial Applications, Graz, Austria\\
$^\ddagger$ Institute of Thermodynamics and Sustainable Propulsion Systems, Graz University of Technology,\\Graz, Austria\\
$^\dagger$ Signal Processing and Speech Communication Laboratory, Graz University of Technology, Graz, Austria
}

\begin{document}
%\ninept
%
\maketitle
\begin{abstract}
It was recently shown that the loss function used for training physics-informed neural networks (PINNs) exhibits local minima at solutions corresponding to fixed points of dynamical systems. 
In the forward setting, where the PINN is trained to solve initial value problems, these local minima can interfere with training and potentially leading to physically incorrect solutions. 
Building on stability theory, this paper proposes a regularization scheme that penalizes solutions corresponding to unstable fixed points. Experimental results on four dynamical systems, including the Lotka-Volterra model and the van der Pol oscillator, show that our scheme helps avoiding physically incorrect solutions and substantially improves the training success rate of PINNs. 
\end{abstract}
\section{Introduction}\label{sec:introduction}

Physics-informed neural networks (PINNs,~\cite{raissi2019physics}) are among the most prominent instantiations of physics-informed machine learning. Capable of including systems of differential equations during training, they have been proposed for solving boundary and initial value problems
% ~\cite{mao2020physics, yu2022gradient}
, for inferring parameters of a differential equation from data
% ~\cite{cai2021physics, chen2020physics},
and for estimating unobservable scalar or vector fields from measurements
% ~\cite{raissi2020hidden}
. 
Especially for the forward problem -- i.e., for learning the solution to a system of differential equations given boundary and initial conditions -- PINNs experience training difficulties for all but the most trivial problem settings. 

%%%%%%%%%%%%%%%%%%%%%%%%%%%%%%%%%%%%%%%%%%%%%%%%%%%%%
%%% Figure: van der Pol
%%%%%%%%%%%%%%%%%%%%%%%%%%%%%%%%%%%%%%%%%%%%%%%%%%%%%
\begin{figure}[t!]
    \centering
    \includegraphics[width=\linewidth]{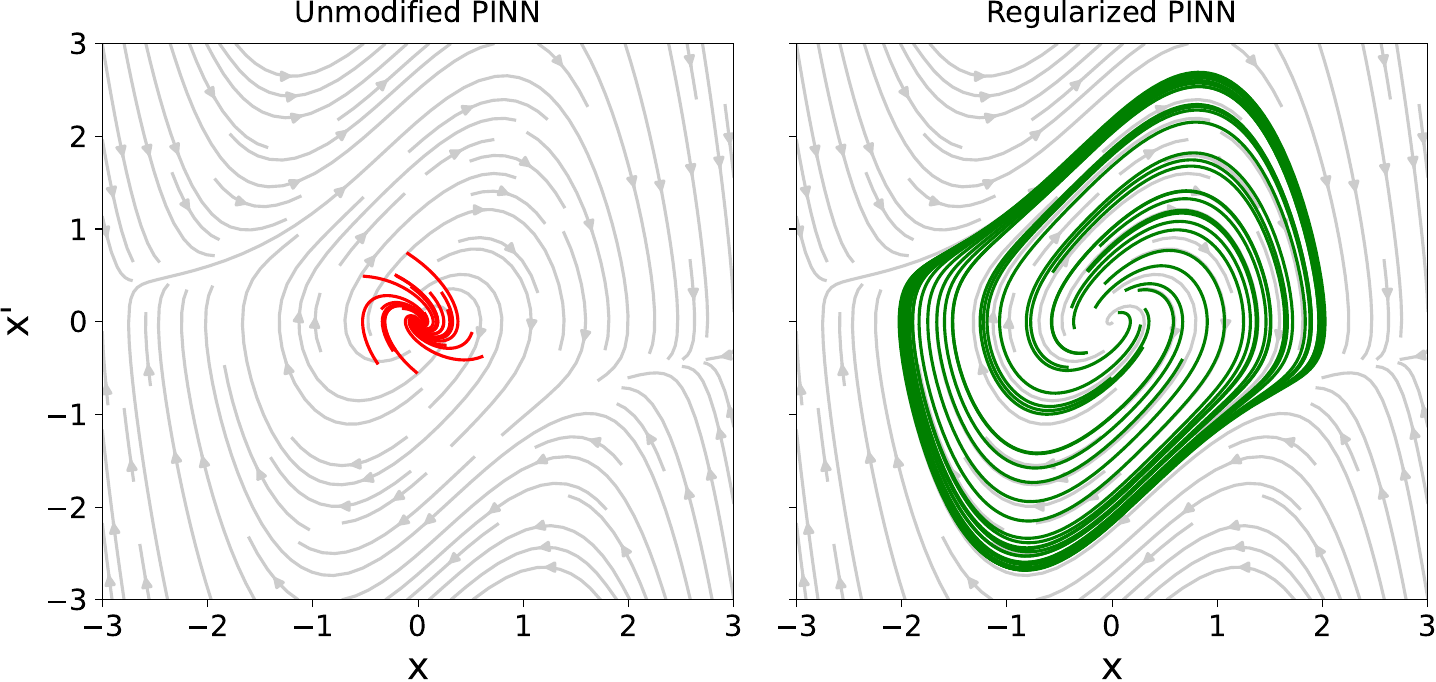}\\
    \includegraphics[width=\linewidth]{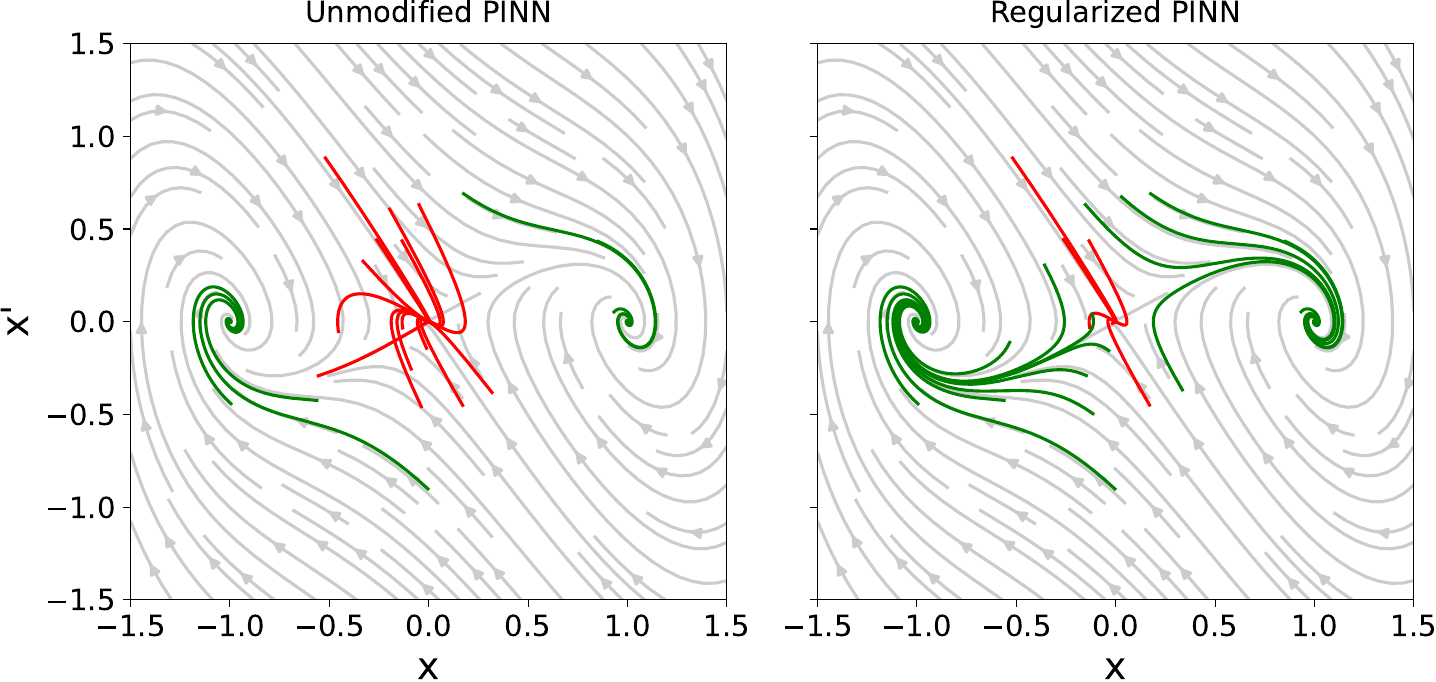}
    \caption{Phase portrait $(x, x')$ of the van der Pol (top) and the unforced Duffing oscillator (bottom) for unmodified (left) and regularized (right) PINN training. Grey arrows indicate the vector field defined by the ODE, with unstable fixed points located at $(x,x')^*=(0,0)$ and stable fixed points (Duffing oscillator) at $(x,x')^*=(\pm1,0)$. Red and green trajectories correspond to unsuccessful and successful PINN solutions. The regularized PINN aligns better with the true dynamics.
    }
    \label{fig:van_der_Pol}
\end{figure}
%%%%%%%%%%%%%%%%%%%%%%%%%%%%%%%%%%%%%

There is a substantial body of literature that investigates the root causes of and offers remedies for these training difficulties. For example, for stiff problems in which the gradient of the solution function varies strongly across the computational domain, it was shown that reducing the weight of points close to gradient maxima leads to a more well-behaved loss and, hence, successful training~\cite{Rohrhofer_FisherPINN_Journal}. 
Other studies attribute challenges in PINN training to the use of large computational domains, where techniques such as domain decomposition~\cite{jagtap2020extended,moseley2023finite} and sinusoidal feature mappings~\cite{Sinusoidal} have been shown to improve training success. 
More generally, PINN training is often susceptible to convergence toward the trivial all-zero solution, as standard neural network initializations tend to bias the model toward this outcome.
To mitigate this, several strategies have been proposed, including specialized initialization schemes~\cite{Sinusoidal}, ensemble methods~\cite{haitsiukevich2023improved}, and adaptive collocation point weighting techniques~\cite{daw2023mitigating,wang2022respecting}.

Interestingly, the all-zero solution is a valid general solution to a large class of differential equations, cf.~\cite[Prop.~1]{Sinusoidal}. In addition to this all-zero solution, the authors of~\cite{rohrhofer2022role} showed that other, non-zero solutions can become minima of the training loss despite corresponding to unphysical behavior. More concretely, in the context of ordinary differential equations (ODEs), it was shown that solutions at fixed points are always global optima of the physics loss, regardless whether these fixed points are stable or not. While convergence to the all-zero solution can be prevented by some of the remedies mentioned above, convergence to unstable, but non-zero fixed points has not been addressed so far.

In this work, we fill this gap by proposing a regularization scheme that avoids training convergence to unstable fixed points, as illustrated in Fig.~\ref{fig:van_der_Pol}. 
More specifically, given a system of ODEs, we linearize the candidate solution around each collocation point and characterize its stability via the eigenvalues of the resulting Jacobian matrix (Section~\ref{sec:methods}). For collocation points for which the candidate solution exhibits unstable behavior, the regularization term is large. Although the approach appears simple, we show in Section~\ref{sec:results} that it substantially improves training success for several (low-dimensional) ODEs (Section~\ref{sec:experimental_setup}).
\iftoggle{arxiv}{}{Due to space limitations, ablation studies, further results, and an in-depth discussion are deferred to an extended version of our manuscript~\cite{Babic_PINNReg_arXiv}.}

% While not shown in this work, we furthermore argue that our regularization can be combined with several other approaches that stabilize PINN training, such as curriculum learning~\cite{krishnapriyan2021characterizing}, domain decomposition~\cite{jagtap2020extended,moseley2023finite}, ensemble approaches~\cite{haitsiukevich2023improved}, or initialization schemes~\cite{Sinusoidal}. We expand on this and other potential extensions of our work (e.g., to partial differential equations) as outlook in Section~\ref{sec:discussion}.

\section{Background}

% \subsection{Dynamical Systems \& Fixed Points}\label{sec:dynamical_systems}
We focus on autonomous dynamical systems that can be described by ODEs of the general form:
\begin{equation}
    x^{(n)} = f(x,x',x'',...,x^{(n-1)}),
\label{eq:ODE}
\end{equation}
where $f$ is a nonlinear function of time $t$ and the unknown solution function $x(t)$ and where $x'$, $x''$, and $x^{(n)}$ denote the first-, second-, and $n$th-order derivative with respect to time $t$. Since any higher-order ODE can be rewritten as a system of first-order ODEs, we consider the multi-dimensional first-order system $x'=f(x)$, where $f=\left(f_1,f_2,...,f_n\right)$ and $x\in\mathbb{R}^n$. For this system, a \emph{fixed point} $x^*\in\mathbb{R}^n$ satisfies $f(x^*)=0$.The local stability around $x^*$ can be determined by linearizing the system through Taylor expansion, and then analyzing the resulting Jacobian matrix 
$J(x^*)= \partial \vec f/\partial \vec x \vert_{x^*}$.
%$J_f=(\partial f_i/\partial x_j)_{i,j=1,..n}$ evaluated at $x^*$. 
If all eigenvalues $\{\lambda_1,...,\lambda_n\}\subset \mathbb{C}$ of $J$ have negative real parts, the fixed point $x^*$ is asymptotically stable~\cite{boyce2004ede}.
If at least one eigenvalue has a positive real part, the fixed point is unstable.
An unstable fixed point is considered a saddle point if it is stable in some directions and unstable in others.

We use PINNs to approximate the unknown solution $x(t)$ to~\eqref{eq:ODE} with a fully connected neural network $x_\theta(t)$, where $\theta$ denotes the network's weights and biases.
The PINN is trained on both labeled and unlabeled data.
Labeled data is used to encode the initial condition (IC) $x(0)$ via the IC loss $\mathcal{L}_{\mathrm{IC}}=\Vert x_\theta(0)-x(0)\Vert^2$.
Unlabeled data (in the context of PINNs also called \textit{collocation points}) is used to enforce the system dynamics~\eqref{eq:ODE} via the physics loss
\begin{equation}
    \mathcal{L}_f=\frac{1}{N_\mathrm{col}}\sum_{i=1}^{N_\mathrm{col}} \Vert  x'_\theta(t_i)-f\left(x_\theta(t_i)\right)\Vert^2,
    \label{eq:physics_loss}
\end{equation}
where the collocation points $\{t_i\}_{i=1}^{N_\mathrm{col}}$ are (randomly) sampled from the temporal domain $0<t\leq T$, where $T$ is the \textit{simulation time}.
Both losses are combined by linear scalarization, i.e. $\mathcal{L}=\mathcal{L}_\mathrm{IC}+\mathcal{L}_f$, where an additional regularization loss, central to our study, will be introduced in Section~\ref{sec:methods}.

As discussed in~\cite{rohrhofer2022role}, fixed points of ODEs correspond to global optima of the physics loss, characterized by non-trivial basins of attraction. 
This arises from the fact that the fixed point condition $f(x^*)=0$ leads to inherently small physics residuals in the vicinity of fixed points. 
Specifically, the physics loss~\eqref{eq:physics_loss} becomes small when $f(x_\theta^\star)\approx0$, i.e. when the network output $x_\theta^\star$ is close to a fixed point $x^*$.
As a consequence, regardless of whether a fixed point is stable or unstable, PINN solutions tend to be attracted to fixed points, often resulting in trajectories that converge to the nearest one (as shown in the left plot of Fig.~\ref{fig:van_der_Pol}). Closely related to convergence issues caused by fixed points is the prominent influence of the trivial zero solution, $x^*=0$, which is a fixed point in many dynamical systems. For instance, Wong et al.~\cite{Sinusoidal} propose learning in sinusoidal space, introducing sinusoidal mappings to initialize PINNs with an appropriate input gradient distribution to overcome trainability issues.

\section{Method} \label{sec:methods}

To prevent the PINN from converging to unstable fixed points, or at least reduce the likelihood of such convergence, we introduce an additional regularization term into the physics loss, designed to modify the loss landscape such that it penalizes predictions near or at unstable fixed points. The proposed regularization term consists of three factors: i) a factor $\mathcal{R}_{\mathrm{LS}}$ that penalizes solutions $x_\theta$ for which the Jacobian indicates local instability at a given collocation point $t_i$; ii) a factor $\mathcal{R}_{\mathrm{SE}}$ that ensures that this regularization is only active at collocation points at which the system has converged to a fixed point, i.e., for which $x'_\theta(t_i)=0$, and iii) a decaying term $C$ that gradually reduces the influence of regularization.

Stability can be easily determined by first converting higher-order ODEs into systems of first-order ODEs and then computing the eigenvalues of the Jacobian matrix obtained through a Taylor expansion around the fixed point. To penalize a candidate solution $x_\theta$ that is unstable at a collocation point $t^*$, we propose the regularization term
\begin{equation}
    \label{eq:reg-unstable-fp}
    \mathcal{R}_{\mathrm{LS}}(t^\star) =  \sum_{\lambda\in\sigma(J(x_\theta(t^\star)))} \max\left( \operatorname{Re}(\lambda), 0\right),
\end{equation}
where $\sigma(J(x_\theta(t^\star)))=\{\lambda_{1}^\star,...,\lambda_{n}^\star\}$ is the spectrum of the Jacobian matrix evaluated at $x_\theta(t^\star)$ and $\operatorname{Re}$ is the real part. 

Since stability is a statement about fixed points, this regularization is not meaningful unless $x'_\theta(t^\star)=0$. We thus multiply $\mathcal{R}_{\mathrm{LS}}(t^\star)$ by a factor $\mathcal{R}_{\mathrm{SE}}(t^\star)$ that is obtained by applying a Gaussian kernel to the time derivatives of the ODE:
\begin{equation}
    \label{eq:Gaussian-kernel}
    \mathcal{R}_{\mathrm{SE}}(t^\star) = \exp \left( -\Vert x'_{\theta}(t^\star)\Vert^2/\varepsilon\right) ,
\end{equation}
where $\varepsilon$ is a hyperparameter that adjusts the sensitivity of the regularization around the fixed point. $\mathcal{R}_{\mathrm{SE}}(t^\star)$  attains its maximum value of one if and only if all time derivatives vanish simultaneously, i.e., if $x_\theta(t^\star)$ is precisely at a fixed point. Candidate solutions converging to unstable fixed points are thus regularized by the term
\begin{equation}
\label{eq:regularization}
    \mathcal{R} = \frac{1}{N}\sum_{i=1}^N \left[ \mathcal{R}_\mathrm{SE}(t_i) \times \mathcal{R}_\mathrm{LS}(t_i) \right],
\end{equation}
where the regularization function is evaluated at the collocation points $\{t_i\}_{i=1}^N$ used to compute the physics loss.

Recognizing the importance of initialization in PINN training, cf.~\cite{Sinusoidal}, it may suffice to steer away the candidate solution $x_\theta$ from unstable fixed points early during training. 
% This suggests an annealing scheme in which the weight of $\mathcal{R}$ decays during training. 
We thus propose a linear decaying scheme that turns off regularization after a certain percentage $\gamma\in(0,1)$ of epochs:
\begin{equation}
    \label{eq:reg-decay}
    C = \max\left(C_0\left(\gamma-\frac{\mathrm{epoch}}{N_\mathrm{epochs}}\right),0\right),
\end{equation}
where $C_0$ denotes the initial regularization coefficient and $N_\mathrm{epochs}$ the total number of epochs. The regularization scheme is finally used in the total loss function via:
\begin{equation}
\label{eq:final_loss}
\mathcal{L}=\mathcal{L}_\mathrm{IC}+\mathcal{L}_f+C\times\mathcal{R}.
\end{equation}

\iftoggle{arxiv}{
The hyperparameters of this scheme are $\varepsilon$ from~\eqref{eq:Gaussian-kernel} as well as $C_0$ and $\gamma$ from the annealing scheme~\eqref{eq:reg-decay}. We show in~\ref{sec:results_sensitivity} that the performance of our approach is not very sensitive to these three hyperparameters.
}{}

\section{Experimental Setup} \label{sec:experimental_setup}
In our analysis, we aim to demonstrate the impact of our regularization on the PINN training performance across four distinct dynamical systems, governed by first- or second-order ODEs in one- or two-dimensional settings.
These systems include the pitchfork bifurcation, unforced Duffing oscillator, van der Pol oscillator, and Lotka-Volterra model.

\textit{PINN Settings.}
The PINNs in our study are fully connected with four hidden layers, each containing $50$ units, and employ the Swish activation function. Training is performed using the Adam optimizer with a learning rate of $0.001$ and a decay rate of $1.0$. During each training epoch, $1024$ collocation points are randomly sampled to compute the physics loss~\eqref{eq:physics_loss}, and the total number of epochs is set to $25k$. 
\iftoggle{arxiv}{%

  \textit{Initial Conditions.}
For first-order ODEs, the ICs are hard-constrained by defining the candidate solution as:
\begin{equation}
    x_\theta(t) = x(0) + \tanh(t) \times \hat{x}_\theta(t),
\end{equation}
where \(\hat{x}_\theta(t)\) represents the network's output. In this case, $\mathcal{L}_\mathrm{IC}$ is not used during PINN training. For second-order ODEs, the ICs are soft-constrained using the loss $\mathcal{L}_\mathrm{IC}$. No weighting scheme is used to balance the IC loss with the physics loss.
}{%
In the case of first-order systems, we use hard constraints for the initial conditions.
}

\textit{Testing Procedure.}
For each system, we train PINNs with and without regularization, for different ICs and simulation times, as convergence to local optima is highly sensitive to these settings, cf.~\cite{rohrhofer2022role}. To account for the inherent randomness in neural network initialization, we train $10$ different networks for each configuration. 
%%%%%%%%%%%%%%%%%%%%%%%%%%%%%%%%%%%%%%%%%%%%%%%%%%%%%
%%% Table: Quantitative Results
%%%%%%%%%%%%%%%%%%%%%%%%%%%%%%%%%%%%%%%%%%%%%%%%%%%%%
\begin{table*}
\centering
\caption{Success rates for unmodified (left number) and regularized (right number) PINN training for different dynamical systems across different simulation times ($T$) and ICs. Bold numbers indicate the best success rate for the considered setting.}
\label{table:quantitative-results}
\footnotesize
% \vspace{0.5cm}
% Row for System 1 Header
\begin{tabular}{p{1.2cm}|p{1.0cm}|p{1.0cm}|p{1.0cm}|p{1.0cm}|p{1.0cm}}
\hline
\multirow{2}{*}{$x(0)$} & \multicolumn{5}{c}{\textbf{Pitchfork Bifurcation}} \\
%\cline{2-6}
& $T=11$ & $T=12$ & $T=13$ & $T=14$ & $T=15$ \\
\hline
\textbf{$(0.1)$} & 0.0 / \textbf{1.0}  & 0.0 / \textbf{1.0} & 0.0 / \textbf{1.0} & 0.0 / \textbf{1.0} & 0.0 / \textbf{1.0} \\
\textbf{$(0.2)$} & 0.0 / \textbf{1.0}  & 0.0 / \textbf{1.0} & 0.0 / \textbf{1.0} & 0.0 / \textbf{1.0} & 0.0 / \textbf{1.0} \\
\textbf{$(0.3)$} & 0.0 / \textbf{1.0}  & 0.0 / \textbf{1.0} & 0.0 / \textbf{1.0} & 0.0 / \textbf{1.0} & 0.0 / \textbf{1.0} \\
\textbf{$(0.4)$} & 0.0 / \textbf{1.0}  & 0.0 / \textbf{1.0} & 0.0 / \textbf{1.0} & 0.0 / \textbf{1.0} & 0.0 / \textbf{1.0} \\
\textbf{$(0.5)$} & 0.1 / \textbf{1.0}  & 0.0 / \textbf{1.0} & 0.1 / \textbf{1.0} & 0.0 / \textbf{1.0} & 0.0 / \textbf{1.0}\\
\hline
\end{tabular}
\vspace{0.1cm}
% Row for System 2 Header
\begin{tabular}{p{1.2cm}|p{1.0cm}|p{1.0cm}|p{1.0cm}|p{1.0cm}|p{1.0cm}}
\hline
\multirow{2}{*}{$x(0), x'(0)$} & \multicolumn{5}{c}{\textbf{Unforced Duffing Oscillator}} \\
%\cline{2-6}
& $T=11$ & $T=12$ & $T=13$ & $T=14$ & $T=15$ \\
\hline
\textbf{$(0.1, 0)$} & \textbf{0.5} / 0.4  & 0.0 / \textbf{0.5} & 0.0 / \textbf{0.7} & 0.0 / \textbf{0.5} & 0.0 / \textbf{0.6} \\
\textbf{$(0.2, 0)$} & \textbf{0.9} / \textbf{0.9}  & 0.1 / \textbf{0.6} & 0.0 / \textbf{0.3} & 0.0 / \textbf{0.5} & 0.0 / \textbf{0.7} \\
\textbf{$(0.3, 0)$} & 0.9 / \textbf{1.0}  & 0.2 / \textbf{0.7} & 0.0 / \textbf{0.3} & 0.0 / \textbf{0.2} & 0.0 / \textbf{0.6} \\
\textbf{$(0.4, 0)$} & \textbf{1.0} / \textbf{1.0}  & 0.2 / \textbf{0.6} & 0.0 / \textbf{0.5} & 0.0 / \textbf{0.5} & 0.0 / \textbf{0.4} \\
\textbf{$(0.5, 0)$} & 0.9 / \textbf{1.0}  & 0.7 / \textbf{0.8} & 0.0 / \textbf{0.4} & 0.0 / \textbf{0.8} & 0.0 / \textbf{0.7} \\
\hline
\end{tabular}
\vspace{0.1cm}
% Row for System 3 Header
\begin{tabular}{p{1.2cm}|p{1.0cm}|p{1.0cm}|p{1.0cm}|p{1.0cm}|p{1.0cm}}
\hline
\multirow{2}{*}{$x(0), x'(0)$} & \multicolumn{5}{c}{\textbf{Van der Pol Oscillator}} \\
%\cline{2-6}
& $T=11$ & $T=12$ & $T=13$ & $T=14$ & $T=15$ \\
\hline
\textbf{$(0.1, 0)$} & 0.0 / \textbf{1.0}  & 0.0 / \textbf{0.2} & 0.0 / \textbf{0.6} & 0.0 / \textbf{0.3} & 0.0 / \textbf{0.1} \\
\textbf{$(0.2, 0)$} & 0.6 / \textbf{1.0}  & 0.0 / \textbf{0.9} & 0.0 / \textbf{0.6} & 0.0 / \textbf{0.5} & 0.0 / \textbf{0.2} \\
\textbf{$(0.3, 0)$} & 0.9 / \textbf{1.0}  & 0.2 / \textbf{1.0} & 0.0 / \textbf{1.0} & 0.0 / \textbf{1.0} & 0.0 / \textbf{0.4} \\
\textbf{$(0.4, 0)$} & \textbf{1.0} / \textbf{1.0}  & 0.8 / \textbf{1.0} & 0.1 / \textbf{0.9} & 0.0 / \textbf{1.0} & 0.0 / \textbf{0.6} \\
\textbf{$(0.5, 0)$} & 0.9 / \textbf{1.0}  & 0.6 / \textbf{1.0} & 0.0 / \textbf{1.0} & 0.0 / \textbf{0.9} & 0.0 / \textbf{0.5} \\
\hline
\end{tabular}
% Row for System 4 Header
\begin{tabular}{p{1.2cm}|p{1.0cm}|p{1.0cm}|p{1.0cm}|p{1.0cm}|p{1.0cm}}
\hline
\multirow{2}{*}{$x(0), y(0)$} & \multicolumn{5}{c}{\textbf{Lotka-Volterra Model}} \\
%\cline{2-6}
 & $T=11$ & $T=12$ & $T=13$ & $T=14$ & $T=15$ \\
\hline
\textbf{$(0.0, 0.1)$} & 0.0 / \textbf{1.0}  & 0.0 / \textbf{1.0} & 0.0 / \textbf{1.0} & 0.0 / \textbf{1.0} & 0.0 / \textbf{1.0} \\
\textbf{$(0.1, 0.0)$} & 0.0 / 0.0  & 0.0 / 0.0 & 0.0 / 0.0 & 0.0 / 0.0 & 0.0 / 0.0 \\
\textbf{$(0.0, 0.5)$} & 0.0 / \textbf{1.0}  & 0.0 / \textbf{1.0} & 0.0 / \textbf{0.9} & 0.0 / \textbf{0.8} & 0.0 / \textbf{0.5} \\
\textbf{$(0.5, 0.0)$} & 0.0 / \textbf{1.0}  & 0.0 / \textbf{1.0} & 0.0 / \textbf{1.0} & 0.0 / \textbf{0.7} & 0.0 / \textbf{0.6} \\
\textbf{$(1.1, 1.1)$} & \textbf{1.0} / \textbf{1.0}  & 0.3 / \textbf{1.0} & 0.0 / \textbf{1.0} & 0.0 / \textbf{1.0} & 0.0 / \textbf{1,0} \\
\hline
\end{tabular}
\vspace{0.1cm}
\end{table*}

\textit{Training Success Rate.}
Our primary goal is to investigate a failure mode where the system converges to a local optimum with fundamentally different dynamics. Thus, rather than focusing on the accuracy of the predictions $x_\theta(t)$, we evaluate training success rates based on whether training converged to an incorrect local optimum or the true solution $x_\mathrm{ref}(t)$.
The reference solution is obtained using the Runge–Kutta method for all systems, except for the pitchfork bifurcation system, for which an analytical solution is used (found in~\cite{rohrhofer2022role}).
A training run is considered successful if the \(L_2\) relative error
\iftoggle{arxiv}{
    \begin{equation}
       \sum_{i=1}^{N} \frac{||x_\theta(t_i) - x_\mathrm{ref}(t_i)||^2}{||x_\mathrm{ref}(t_i)||^2},
    \end{equation}
    remains below $0.15$, where $\{t_i\}_{i=1}^N$ are sampled equidistantly from $[0,T]$ with a spacing of 0.01.
}{from the reference solution remains below $0.15$.}

\textit{Hyperparameters.}
For the training with regularization term added to the loss, we set the hyperparameter $\varepsilon$ to $0.01$, initial regularization coefficient $C_0$ to $1.0$ and the fraction of epochs $\gamma$ to $0.5$. We study the sensitivity of these hyperparameters in Section~\ref{sec:results_sensitivity}.

\textit{Candidate Systems:}
% \subsection{Pitchfork Bifurcation}\label{sec:pitchfork}
%
The \textbf{pitchfork bifurcation} system can be used to model population dynamics with negative growth effects at both low and high population levels. 
It is described by the first-order ODE $x'=x-x^3$ and has one unstable fixed point at $x^* = 0$ and two (asymptotically) stable fixed points at $x^* = 1$ and $x^* = -1$. 
As shown in~\cite{rohrhofer2022role}, the unstable fixed point can cause severe convergence issues for PINNs when the simulation time is long and/or the ICs are close to the unstable point.%
% \subsection{Unforced Duffing Oscillator}\label{sec:duffing}
% We tested the effects of the regularization scheme on the second-order ODE representing the unforced damped Duffing oscillator with the following form:
The \textbf{unforced damped Duffing oscillator} is described by the second-order ODE $x'' + x' - x + x^3 = 0$, which gives an unstable saddle point at $(x,x')^*=(0,0)$ and two (asymptotically) stable fixed points at $(x,x')^*=(\pm1,0)$. %
% \subsection{Van der Pol System}\label{sec:van_der_pol}
 % 
 The \textbf{Van der Pol oscillator} \cite{vanderPol01111926} is a dynamical system described by the second-order ODE of the form $x'' - (1 - x^2)x' + x = 0$.
The system exhibits stable oscillations, referred to in the literature as relaxation oscillations, and possesses a single unstable fixed point at $(x,x')^*=(0,0)$.%
% \subsection{Lotka-Volterra Model}\label{sec:lotka_volterra}
% 
The \textbf{Lotka–Volterra equations} are a well-known system of first-order ODEs used to describe predator–prey dynamics in biological systems.
While the system's behavior can be complex depending on its specific formulation, in our work we consider a generalized version of the Lotka-Volterra equations:
\begin{equation} 
\label{eq:2d-toy-example-ODE} 
\begin{split} 
x'&= x(3 - x - 2y), \\
y'&= y(2 - x - y).
\end{split} 
\end{equation}
This system exhibits two (asymptotically) stable fixed points at $(x, y)^* = (0, 2)$ and $(x, y)^* = (3, 0)$, and two unstable fixed points at $(x, y)^* = (0, 0)$ and $(x, y)^* = (1, 1)$, with the latter being a saddle point.

\section{Results}\label{sec:results}
\textit{Training Success Rates.} Table \ref{table:quantitative-results} shows the quantitative results of our experiments for unmodified PINN training and for our proposed regularization scheme~\eqref{eq:final_loss} with default hyperparameters as reported in Section~\ref{sec:experimental_setup}. We deliberately selected ICs and simulation times such that unmodified PINN training exhibits convergence problems (e.g., the simulation times were chosen longer for the pitchfork bifurcation system than in~\cite{rohrhofer2022role}). Furthermore, we did not apply training modifications proposed previously, such as loss weighting, adaptive collocation point sampling, etc., as they are orthogonal to our regularization scheme. In all of the unsuccessful training runs, training converged to the local optimum corresponding to the unstable fixed point of the dynamical system, confirming the results of~\cite{rohrhofer2022role} that unstable fixed points play a major role in PINN training. As the table shows, regularization substantially improves success rates for all considered systems and problem parameterizations, validating the effectiveness of our proposal even for longer simulation times.

%%%%%%%%%%%%%%%%%%%%%%%%%%%%%%%%%%%%%%%%%%%%%%%%%%%%%

\textit{Unforced Duffing and van der Pol Oscillators.}
For the unforced Duffing and the van der Pol oscillator we present an additional evaluation. Specifically, we fix the simulation time at $T=12.5$ and randomly sample 20 ICs from a Gaussian distribution centered at the unstable fixed point $(x,x')^*=(0,0)$, with a covariance matrix $\Sigma=0.25\mathbb{I}$.
The simulated trajectories in the phase portrait are shown in Fig.~\ref{fig:van_der_Pol}. Consistent with the results in Table~\ref{table:quantitative-results}, the unmodified PINN frequently fails by converging to the unstable fixed point, whereas the regularized PINN exhibits much better performance. Indeed, the training success rates are up from 30\% to 70\% for the Duffing and from 0\% to 100\%  for the van der Pol oscillator.

\subsection{Hyperparameter Sensitivity Analysis}\label{sec:results_sensitivity}
We analyze the sensitivity of our regularization scheme w.r.t.\ the three hyperparameters $C_0$, $\varepsilon$ and $\lambda$. Rather than varying all three hyperparameters simultaneously, which would be computationally expensive, in this experiment we keep two parameters fixed and vary the third. \\
\indent We conduct the experiment on the Duffing oscillator using the IC $(x_0, x_0') = (0.01, 0.0)$ and simulation time of $T=15$. 
While we believe that the optimal hyperparameters still depend on the considered system, its ICs, and the required simulation time, the results presented in Table~\ref{tab:sensitivity} suggests that the performance of our regularization scheme depends only mildly on even large variations of the hyperparameters, and always improves upon unmodified PINN training (which in this case has a success rate of 0\% for the considered 20 runs).

\begin{table}[h]
\centering
\caption{Hyperparameter sensitivity analysis for the proposed regularization scheme. Success rates are shown for 20 training runs for the Duffing oscillator with IC $(x_0, x_0') = (0.01, 0.0)$ and simulation time $T=15$. Unmodified PINN training fails for this setting.}
\label{tab:sensitivity}
\footnotesize
% Subtable 1: Varying C
\begin{tabular}{l|c|c|c|c|c|c|c}
\multicolumn{8}{c}{$\epsilon=0.01$,  $\gamma=0.5$}\\\hline
\textbf{$C$} & 0.001 & 0.01 & 0.1 & 1.0 & 10.0 & 100.0 & 1000.0 \\ \hline
\textbf{Success} & 0 & 0.3 & 0.4 & 0.35 & 0.6 & 0.6 & 0.6 \\ \hline
\end{tabular}
\vspace{.3cm}

% Subtable 2: Varying eps
\begin{tabular}{l|c|c|c|c|c|c|c}
\multicolumn{8}{c}{$C=1.0$,  $\gamma=0.5$}\\\hline
\textbf{$\epsilon$}&0.0001 & 0.001 & 0.01 & 0.1 & 1.0 & 10.0 & 100 \\ \hline
\textbf{Success} & 0.5 & 0.4 & 0.45 & 0.45 & 0.35 & 0.6 & 0.4 \\ \hline
\end{tabular} 
\vspace{.3cm}

% Subtable 3: Varying frac
\begin{tabular}{l|c|c|c|c|c|c|c|c|c}
\multicolumn{10}{c}{$\epsilon=0.01$, $C=1.0$}\\\hline
\textbf{$\gamma$} &0.1 & 0.2 & 0.3 & 0.4 & 0.5 & 0.6 & 0.7 & 0.8 & 0.9 \\ \hline
\textbf{Success} & 0.5 & 0.7 & 0.45 & 0.4 & 0.7 & 0.75 & 0.6 & 0.6 & 0.4 \\ \hline
\end{tabular}
%\vspace{.1cm}
\end{table}

\section{Conclusion}
In this work, we introduced a regularization scheme for avoiding unstable fixed points when training PINNs on dynamical systems governed by ODEs. This approach proves especially useful in challenging scenarios, such as those involving long simulation times or initial conditions near unstable fixed points, where training would otherwise be difficult or infeasible. While this method does not entirely eliminate convergence to local optima, it can be effectively combined with other strategies designed to mitigate such failure modes. 

Although ODEs remain relevant in many fields, including control theory, systems biology, and mechanical engineering, the extension of this approach to partial differential equations (PDEs) is a compelling direction for future work. This is not a trivial step, as analyzing the stability of steady-state solutions in PDEs is significantly more complex. Nonetheless, our results demonstrate that explicitly addressing unstable fixed points can substantially improve PINN training, providing a promising foundation for broader applications.

\iftoggle{arxiv}{}{\clearpage}
\section*{Acknowledgments}
This work was supported by the European Union’s HORIZON Research and Innovation Programme under grant agreement No 101120657, project ENFIELD (European Lighthouse to Manifest Trustworthy and Green AI).
%%%%%%%%%%%%%%%%%%%%%%%%%%%%%%%%%%%%%%%%%%%%%%%%%%%%%%%%%%%%%%%%%%%%%%%%

%%% Use this command to include your bibliography file.

\bibliographystyle{IEEEbib}
\bibliography{mybibfile}

\begin{thebibliography}{10}

\bibitem{raissi2019physics}
Maziar Raissi, Paris Perdikaris, and George~E Karniadakis,
\newblock ``Physics-informed neural networks: A deep learning framework for solving forward and inverse problems involving nonlinear partial differential equations,''
\newblock {\em Journal of Computational Physics}, vol. 378, pp. 686--707, 2019.

\bibitem{Rohrhofer_FisherPINN_Journal}
Franz~M. Rohrhofer, Stefan Posch, Clemens G\"o\ss{}nitzer, and Bernhard~C. Geiger,
\newblock ``Approximating families of sharp solutions to {F}isher's equation with physics-informed neural networks,''
\newblock {\em Computer Physics Communications}, vol. 307, pp. 109422, 2025,
\newblock open-access: {\tt arXiv:2402.08313 [cs.LG]}.

\bibitem{jagtap2020extended}
Ameya~D Jagtap and George~Em Karniadakis,
\newblock ``Extended physics-informed neural networks {(XPINNs)}: A generalized space-time domain decomposition based deep learning framework for nonlinear partial differential equations,''
\newblock {\em Communications in Computational Physics}, vol. 28, no. 5, 2020.

\bibitem{moseley2023finite}
Ben Moseley, Andrew Markham, and Tarje Nissen-Meyer,
\newblock ``Finite basis physics-informed neural networks {(FBPINNs)}: a scalable domain decomposition approach for solving differential equations,''
\newblock {\em Advances in Computational Mathematics}, vol. 49, no. 4, pp. 62, 2023.

\bibitem{Sinusoidal}
Jian~Cheng Wong, Chin~Chun Ooi, Abhishek Gupta, and Yew-Soon Ong,
\newblock ``Learning in sinusoidal spaces with physics-informed neural networks,''
\newblock {\em IEEE Transactions on Artificial Intelligence}, vol. 5, no. 3, pp. 985--1000, 2024.

\bibitem{haitsiukevich2023improved}
Katsiaryna Haitsiukevich and Alexander Ilin,
\newblock ``Improved training of physics-informed neural networks with model ensembles,''
\newblock in {\em Proc. Int. Joint Conf. on Neural Networks (IJCNN)}, 2023, pp. 1--8.

\bibitem{daw2023mitigating}
Arka Daw, Jie Bu, Sifan Wang, Paris Perdikaris, and Anuj Karpatne,
\newblock ``Mitigating propagation failures in physics-informed neural networks using retain-resample-release {(R3)} sampling,''
\newblock in {\em Proc. Int. Conf. on Machine Learning (ICML)}, Honolulu, Hawaii, USA, July 2023, pp. 7264--7302.

\bibitem{wang2022respecting}
Sifan Wang, Shyam Sankaran, and Paris Perdikaris,
\newblock ``Respecting causality for training physics-informed neural networks,''
\newblock {\em Computer Methods in Applied Mechanics and Engineering}, vol. 421, pp. 116813, 2024.

\bibitem{rohrhofer2022role}
Franz~M. Rohrhofer, Stefan Posch, Clemens G\"o\ss{}nitzer, and Bernhard~C. Geiger,
\newblock ``On the role of fixed points of dynamical systems in training physics-informed neural networks,''
\newblock {\em Trans. Machine Learning Research}, , no. 1, 2023,
\newblock Open-access.

\bibitem{Babic_PINNReg_arXiv}
Milos Babic, Franz~M. Rohrhofer, and Bernhard~C. Geiger,
\newblock ``Stabilizing {PINNs}: A regularization scheme for {PINN} training to avoid unstable fixed points of dynamical systems,'' 2025,
\newblock \texttt{arXiv:25XX.XXXXX}.

\bibitem{boyce2004ede}
W.E. Boyce and R.C. DiPrima,
\newblock {\em {Elementary differential equations and boundary value problems}},
\newblock Wiley New York, 8th edition, 2004.

\bibitem{vanderPol01111926}
Balth. van~der Pol~Jun.,
\newblock ``{LXXXVIII}. on “relaxation-oscillations”,''
\newblock {\em The London, Edinburgh, and Dublin Philosophical Magazine and Journal of Science}, vol. 2, no. 11, pp. 978--992, 1926.

\bibitem{krishnapriyan2021characterizing}
Aditi Krishnapriyan, Amir Gholami, Shandian Zhe, Robert Kirby, and Michael~W Mahoney,
\newblock ``Characterizing possible failure modes in physics-informed neural networks,''
\newblock {\em Advances in neural information processing systems}, vol. 34, pp. 26548--26560, 2021.

\bibitem{wang2024respecting}
Sifan Wang, Shyam Sankaran, and Paris Perdikaris,
\newblock ``Respecting causality for training physics-informed neural networks,''
\newblock {\em Computer Methods in Applied Mechanics and Engineering}, vol. 421, pp. 116813, 2024.

\bibitem{mojgani2022lagrangian}
Rambod Mojgani, Maciej Balajewicz, and Pedram Hassanzadeh,
\newblock ``Lagrangian {PINNs}: A causality-conforming solution to failure modes of physics-informed neural networks,''
\newblock {\em arXiv preprint arXiv:2205.02902}, 2022.

\bibitem{guo2022novel}
Jia Guo, Haifeng Wang, and Chenping Hou,
\newblock ``A novel adaptive causal sampling method for physics-informed neural networks,''
\newblock {\em arXiv preprint arXiv:2210.12914}, 2022.

\end{thebibliography}

\iftoggle{arxiv}{
\appendix
\section{Additional Results}

%%%%%%%%%%%%%%%%%%%%%%%%%%%%%%%%%%%%%%%%%%%%%%%%%%%%%
%%% Figure: Pitchfork Bifuraction
%%%%%%%%%%%%%%%%%%%%%%%%%%%%%%%%%%%%%%%%%%%%%%%%%%%%%
\begin{figure}[t]
    \centering  
    \includegraphics[width=\columnwidth]{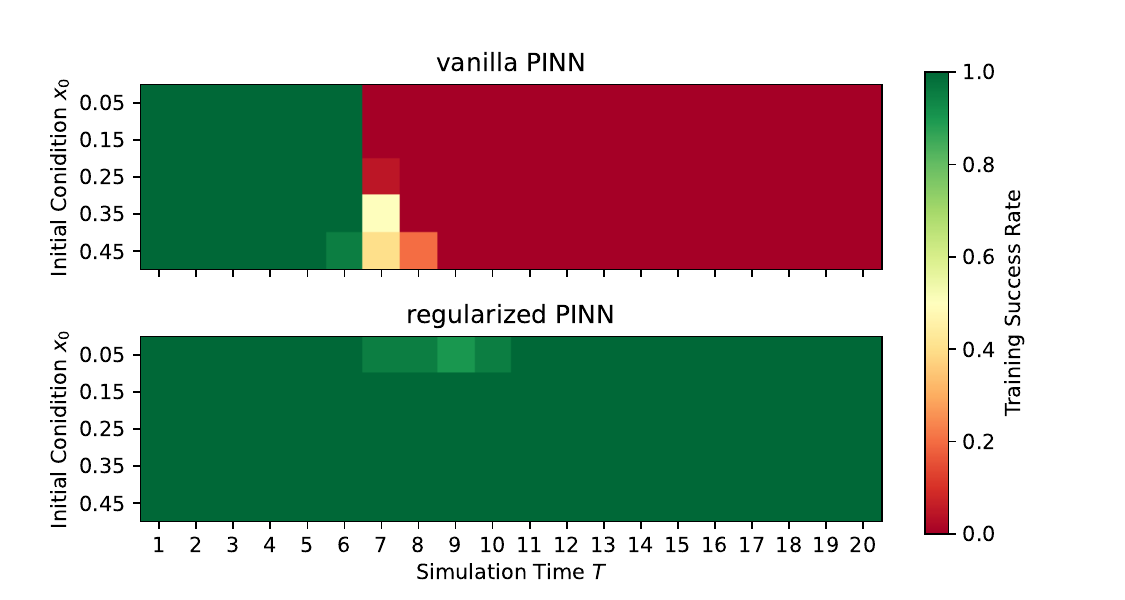}
    \caption{Success rates of unmodified (top) and regularized (bottom) training runs over different initial conditions $x_0$ and simulation times $T$.}
    \label{fig:pitchfork_bifucation}
    \vspace{1cm}
\end{figure}
%%%%%%%%%%%%%%%%%%%%%%%%%%%%%%%%%%%%%%%%%%%%%%%%%%%%%

\textit{Pitchfork Bifurcation.}
As shown in Table~\ref{table:quantitative-results}, the regularized PINN achieves perfect training success rates, while the unmodified PINN nearly completely fails in every tested setting chosen for the study.
We therefore extend the numerical study by selecting simulation times from a broader range. Specifically, we repeat the experiment for simulation times $T\in\{1,…,20\}$ to better capture the regions where the unmodified PINN training still succeeds. %and the regularized PINN training begins to fail.
The results are visualized as a heatmap in Fig.~\ref{fig:pitchfork_bifucation}. From the figure, we can observe that while the unmodified PINN training begins to fail at approximately $T=7$, the regularized PINN continues to achieve successful training for all of the tested simulation times.
Only for simulation times in the range $7 \ge T \ge 10$ with an IC of $x_0 = 0.05$, the regularized PINN exhibits a slight performance degradation, which we believe to be an artifact of the specific experimental settings and potentially resolvable through more optimized hyperparameter tuning.

\textit{Lotka-Volterra Model.}
For the Lotka–Volterra model, we uniformly sample 50 ICs from the interval $(-1, 4)^2$ along both x and y axis, and use a fixed simulation time of $T=12.5$. 
The phase portrait for this system is presented in Fig.~\ref{fig:lotka_volterra}. While the proposed regularization is not as effective as in the two second-order ODEs (cf. Figs.~\ref{fig:van_der_Pol}), it still yields a notable improvement in training success rate from 64\% to 74\%.
%: 37 out of 50 training runs were successful with regularization, compared to 32 without.

%%%%%%%%%%%%%%%%%%%%%%%%%%%%%%%%%%%%%%%%%%%%%%%%%%%%%
%%% Figure: Lotka-Volterra
%%%%%%%%%%%%%%%%%%%%%%%%%%%%%%%%%%%%%%%%%%%%%%%%%%%%%
\begin{figure}[t]
    \centering  
    \includegraphics[width=\columnwidth]{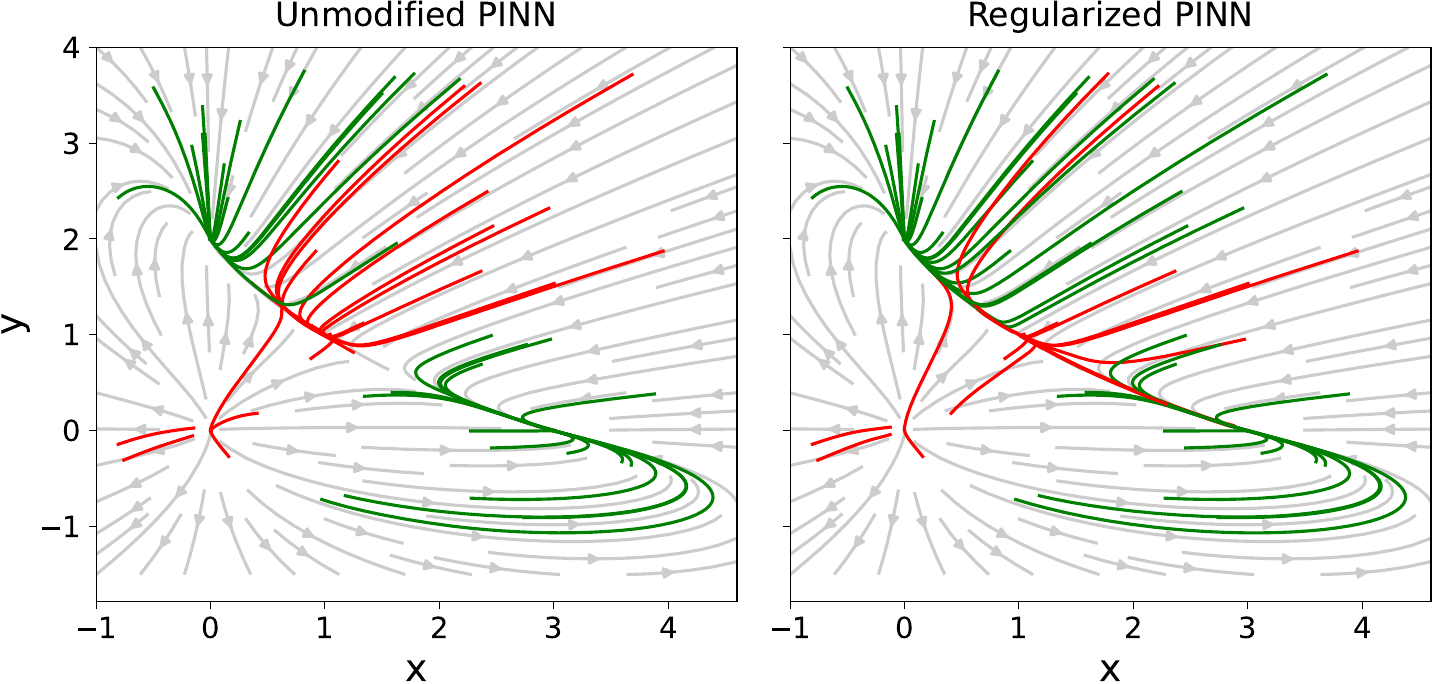}
    \caption{Phase portrait $(x, y)$ of the Lotka-Volterra for unmodified (left) and regularized (right) PINN training. Grey arrows indicate the vector field defined by the ODE, with an unstable fixed point located at $(x,y)^*=(0,0)$ and a saddle points located at $(x,y)^*=(1,1)$. Red trajectories correspond to unsuccessful, and green trajectories to successful PINN training outcomes. While the proposed regularization is not as effective as for the other systems, it still yields a notable improvement in training success rates.}
    \label{fig:lotka_volterra}
    \vspace{1cm}
\end{figure}
%%%%%%%%%%%%%%%%%%%%%%%%%%%%%%%%%%%%%%%%%%%%%%%%%%%%%

\subsection{Ablation Study}
Penalizing candidate solutions that converge to static equilibria via~\eqref{eq:Gaussian-kernel} is, when combined with the coefficient decay~\eqref{eq:reg-decay}, a valid regularization scheme on its own -- at least for systems that do not have (asymptotically) stable fixed points. In this section, we thus investigate the performance of PINN training according to~\eqref{eq:final_loss}, both with and without including the local stability term~\eqref{eq:reg-unstable-fp} in the rgularization loss. 

To this effect, we compare the two schemes on the Duffing oscillator, as its physical trajectories converges to one of the asymptotically stable fixed points, a property for which taking stability into account seems as the most sensible approach. We also compared the two schemes for the Van der Pol oscillator, a system for which the regularization scheme shown in \eqref{eq:Gaussian-kernel} is, in theory, a valid scheme on its own as the system does not exhibit any stable fixed point. 
We perform the experiment on a single simulation time and IC, but optimized the PINN performance over all three hyperparameters in a grid search for each regularization scheme on each system separately. We chose a simulation time of $T=15$ for both systems and ICs $(x_0, x'_0) = (0.01, 0.0)$ for the Duffing oscillator and $(x_0, x'_0) = (0.1, 0.0)$ for the Van der Pol oscillator. These settings were deliberately chosen to avoid perfect success rates for either approach, thereby creating challenging conditions that highlight meaningful differences between the two methods.
Table~\ref{table:ablation-study} reports the highest success rates achieved during hyperparameter optimization. As shown, the best performance achieved by the combined loss function~\eqref{eq:final_loss} is either comparable to (in the case of the van der Pol oscillator) or superior to (in the case of the unforced Duffing oscillator) that achieved by the static equilibrium regularization~\eqref{eq:Gaussian-kernel} alone. 
(There may not be an ordering between the two regularization schemes for fixed sets of hyperparameters, however.) Thus, we conclude that regularizing against unstable fixed points is necessary for systems that have asymptotically stable fixed points, and does not deteriorate performance for systems that do not.

%%%%%%%%%%%%%%%%%%%%%%%%%%%%%%%%%%%%%%%%%%%%%%%%%%%%%
%%% Table: Ablation Study
%%%%%%%%%%%%%%%%%%%%%%%%%%%%%%%%%%%%%%%%%%%%%%%%%%%%%
\begin{table}[h]
\centering
\caption{Comparing regularization with and without taking local stability $\mathcal{L}_\mathrm{LS}$ into account. The success rate computed over 20 training runs, maximized over all hyperparameter combinations, is shown.}
\label{table:ablation-study}
\small
\begin{tabular}{p{3cm} l c}
\hline
\textbf{Dynamical System} & \textbf{Regularization} & \textbf{Success Rate} \\
\hline
\multirow{3}{=}{Unforced Duffing Oscillator\\ ($T=15$, $x_0=0.01$)} 
  & -- & 0 \\
  & $\mathcal{R}_\mathrm{SE}$ & 0.6 \\
  & $\mathcal{R}_\mathrm{SE} \times \mathcal{R}_\mathrm{LS}$ & 0.7 \\
\hline
\multirow{3}{=}{Van der Pol Oscillator ($T=15$, $x_0=0.1$)} 
  & -- & 0 \\
  & $\mathcal{R}_\mathrm{SE}$ & 0.7 \\
  & $\mathcal{R}_\mathrm{SE} \times \mathcal{R}_\mathrm{LS}$ & 0.7 \\
\hline
\end{tabular}
\end{table}
%%%%%%%%%%%%%%%%%%%%%%%%%%%%%%%%%%%%%%%%%%%%%%%%%%%%%

\section{Discussion, Limitation \& Outlook} \label{sec:discussion}

\textit{Training Difficulties due to Fixed Points.}
In~\cite{rohrhofer2022role}, it was shown that fixed points—regardless of their local stability—correspond to local minima in the physics loss function, which can lead to severe convergence issues in PINNs, particularly for long simulation times and/or when ICs are close to these fixed points.
Closely related are common training difficulties in PINNs caused by the zero solution $x^*=0$, which is a trivial solution and fixed point for many dynamical systems. This solution is especially attractive due to common network initialization schemes that tend to bias the model toward it.
Our experiments on the Lotka–Volterra model demonstrated that fixed points may cause convergence issues not only when the IC is close to them, but also when trajectories pass near them—even if the IC is set far away.
This specific scenario is illustrated in Fig.~\ref{fig:lotka_volterra}, where simulated trajectories are drawn toward the unstable fixed point at $(x, y)^* = (1, 1)$, despite originating from regions that are initially far away.

\textit{Overall Performance.} Our experiments consistently highlight the prominent role of fixed points in training PINNs, demonstrating that the vanilla, unmodified PINN framework struggles to simulate dynamical systems in which fixed points interfere—either because trajectories originate near them or pass close to them during the simulation.
With our proposed regularization scheme, introduced in Section~\ref{sec:methods}, we specifically aim to mitigate these training difficulties by leveraging both the fixed point condition, which states that $f(x^*)=0$, and the local stability information provided by the eigenvalues $\{\lambda_i\}$ of the Jacobian $J(x^*)$.
Our results on various dynamical systems demonstrate that incorporating the regularization scheme into the overall loss function substantially improves training success rates. 
This improvement was shown quantitatively in Table~\ref{table:quantitative-results} and qualitatively in Figs~\ref{fig:van_der_Pol}, \ref{fig:pitchfork_bifucation}, and \ref{fig:lotka_volterra}.

\textit{Limitations due to Saddle Points.} Although the training success rate improved across many tested settings, our results also indicate that the proposed regularization scheme does not fully eliminate training difficulties.
Manual inspection of the training process (not shown in the manuscript) revealed that many unsuccessful training outcomes—despite the proposed regularization being active—can be traced to the presence of saddle points, which constitute a specific class of unstable fixed points. Saddle points are characterized by the coexistence of stable and unstable manifolds: along certain directions (corresponding to negative eigenvalues of the Jacobian), trajectories are attracted toward the fixed point, while along others (positive eigenvalues), they are repelled (cf. Figs~\ref{fig:van_der_Pol} and \ref{fig:lotka_volterra}).
This mixed stability behavior leads to a critical sensitivity to ICs. In particular, the stable manifold of a saddle point effectively acts as a \textit{separatrix}, dividing the phase space into regions of qualitatively different dynamics. 
If the candidate solution $x_\theta(t)$ at network  initialization  lies on the “wrong side” of this separatrix, the regularization term—which is designed to repel trajectories from fixed points—may inadvertently drive the solution further into the repelling region. Once the regularization decays or is deactivated, the dynamics governed solely by the physical loss may guide the trajectory back toward the saddle point, ultimately resulting in convergence failure.
In contrast, if the initial candidate solution lies on the “correct side” of the stable manifold, the regularization successfully steers the trajectory away from the saddle point, allowing it to follow the correct streamlines toward the physically meaningful solution. 
This limitation highlights the importance of network initialization in the presence of saddle-type fixed points, where the delicate geometry of the phase space can significantly affect training outcomes.

\textit{Outlook.}
Recognizing this limitation of the proposed regularization scheme motivates the discussion of combining it with existing techniques from the literature. For instance, sequence-to-sequence learning~\cite{krishnapriyan2021characterizing} or causality-respecting methods~\cite{wang2024respecting,mojgani2022lagrangian,guo2022novel} can be naturally integrated with our approach to progressively extend the region of the simulation. These methods allow the training process to begin in a narrow temporal window around the IC and gradually expand toward the full simulation domain as training progresses.
This approach could potentially circumvent the limitations introduced by saddle points, as their separatrices would not significantly influence the optimization during the initial training phase.

Finally, we would like to comment on the role of the static equilibrium regularization term~\eqref{eq:Gaussian-kernel} within the overall regularization framework.
This component is designed to prevent PINN training from becoming trapped in trivial solutions that remain constant over time.
Although its effectiveness was demonstrated only for systems governed by ODEs, we believe that it is also applicable to dynamical systems described by partial differential equations.
Given that PINN training has also been shown to be adversely affected by steady-state solutions in PDE-based systems~\cite{rohrhofer2022role}, the static equilibrium regularization may help push the solution away from such steady states during the early stages of training, thereby facilitating convergence toward transient, physically meaningful dynamics.

}
{}

\end{document}